\documentclass[letterpaper, 10 pt, conference]{ieeeconf}  

\IEEEoverridecommandlockouts                              

\usepackage{graphics} 
\usepackage{epsfig} 
\usepackage{color}
\usepackage[dvipsnames]{xcolor}
\usepackage{algorithm}
\usepackage{algorithmic}
\usepackage{todonotes}
\usepackage{graphicx}
\usepackage{amsmath,amssymb}
\usepackage{subfigure}
\usepackage{placeins}
\usepackage{booktabs}
\usepackage{balance}
\usepackage{hyperref}

\newcommand\bbE{\ensuremath{\mathbb{E}}} 
\DeclareMathOperator*{\expectation}{\bbE}

\title{\LARGE \bf
Learning from Demonstration in the Wild
}

\author{
  Feryal Behbahani$^{1}$, Kyriacos Shiarlis$^{1}$, Xi Chen$^{1}$, Vitaly Kurin$^{1, 2}$, Sudhanshu Kasewa$^{1, 2}$, Ciprian Stirbu$^{1, 2}$,\\Jo\~{a}o Gomes$^{1}$, Supratik Paul$^{1, 2}$, Frans A.~Oliehoek$^{1, 3}$, Jo\~{a}o Messias$^{1}$, Shimon Whiteson$^{1, 2}$\\
  \thanks{$^{1}$Latent Logic, Oxford, England}
\thanks{$^{2}$University of Oxford, Oxford, England
}
\thanks{$^{3}$Delft University of Technology, Delft, Netherlands
}%
}
\begin{document}

\maketitle
\thispagestyle{empty}
\pagestyle{empty}

\begin{abstract}
\emph{Learning from demonstration} (LfD) is useful in settings where hand-coding behaviour or a reward function is impractical.
It has succeeded in a wide range of problems but typically relies on manually generated demonstrations or specially deployed sensors and has not generally been able to leverage the copious demonstrations available \emph{in the wild}:
those that capture behaviours that were occurring anyway using sensors that were already deployed for another purpose, e.g., traffic camera footage capturing demonstrations of natural behaviour of vehicles, cyclists, and pedestrians.
We propose \emph{video to behaviour} (ViBe), a new approach to learn models of behaviour from unlabelled raw video data of a traffic scene collected from a single, monocular, initially uncalibrated camera with ordinary resolution.
Our approach calibrates the camera, detects relevant objects, tracks them through time, and uses the resulting trajectories to perform LfD, yielding models of naturalistic behaviour.
We apply ViBe to raw videos of a traffic intersection and show that it can learn purely from videos, without additional expert knowledge.
\end{abstract}

\section{Introduction}
\label{intro}
\emph{Learning from demonstration} (LfD) is a machine learning technique that can learn complex behaviours from a dataset of expert trajectories, called \textit{demonstrations}.
LfD is particularly useful in settings where hand-coding behaviour, or engineering a suitable reward function, is too difficult or labour intensive.
While LfD has succeeded in a wide range of problems \cite{zhang2017deep, osa2018algorithmic, argall2009survey}, nearly all methods rely on either artificially generated demonstrations
(e.g., in laboratory settings) or those collected by specially deployed sensors (e.g., MOCAP).
These restrictions greatly limit the practical applicability of LfD, which to date has largely not been able to leverage the copious demonstrations available
\emph{in the wild}: those that capture behaviour that was occurring anyway using sensors that were already deployed for other purposes.

For example, consider the problem of training autonomous vehicles to navigate in the presence of human road users.
Since physical road tests are expensive and dangerous, simulation is an essential part of the training process.
However, such training requires a realistic simulator which, in turn, requires realistic models of other agents, e.g., vehicles, cyclists, and pedestrians, that the autonomous vehicle interacts with.
Hand-coded models of road users are labour intensive to create, do not generalise to new settings, and do not capture the diversity of behaviours produced by humans.

\begin{figure}
  \centering
  \includegraphics[width=0.4\textwidth]{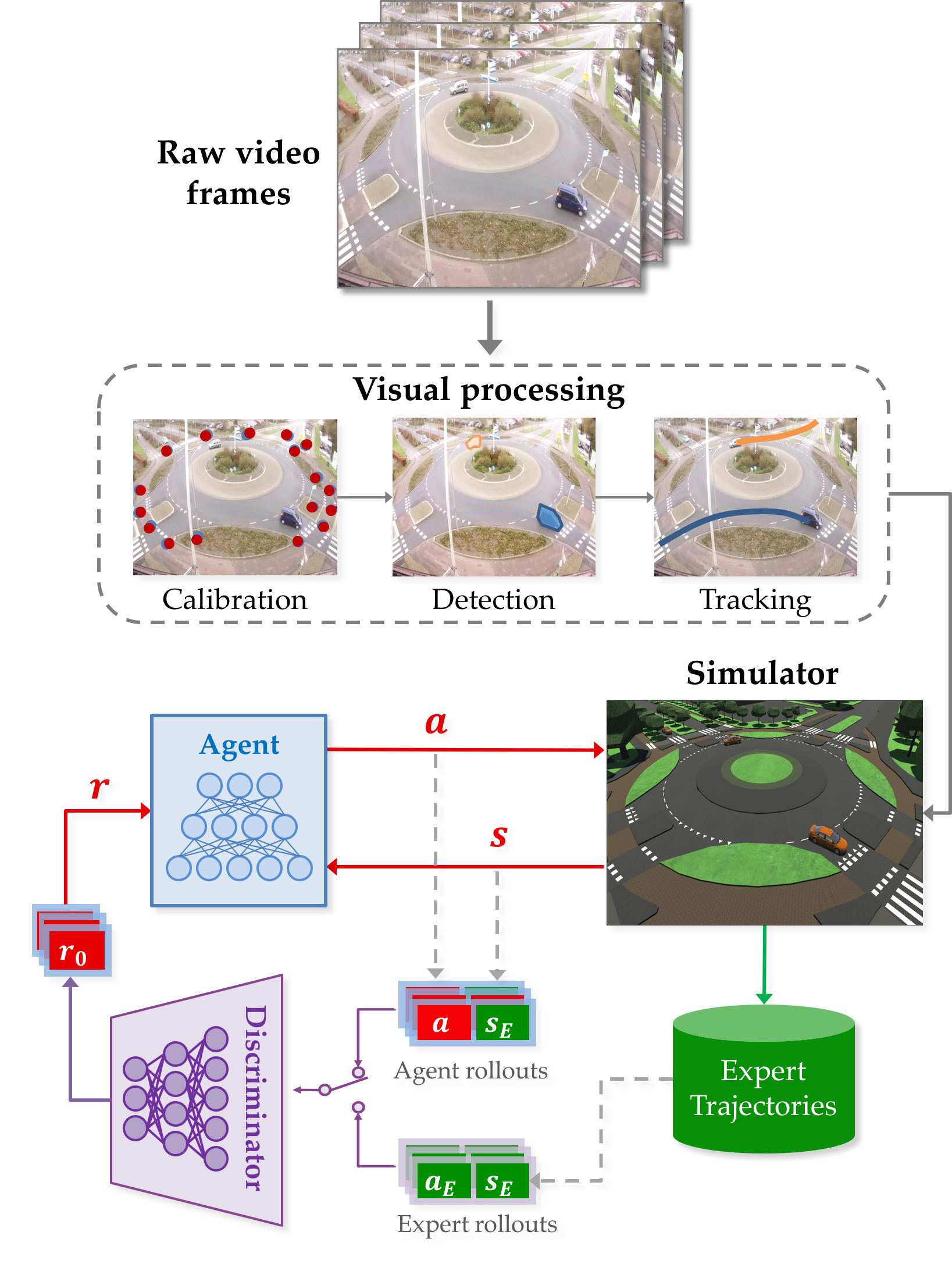}
  \caption{Schematic of the ViBe approach}
  \label{fig:vibe}
  \vspace{-13pt}

\end{figure}

LfD is an attractive alternative. In principle, subjects could be recruited to
demonstrate such behaviour or existing road users could be augmented with
sensors to record their trajectories. However, doing so would be expensive and
yield only limited datasets. A more effective way would be to use the abundance
of relevant demonstrations available in the wild, such as traffic camera footage. Unfortunately, there are currently no LfD methods that can
learn from such sources of traffic demonstrations.

In this paper, we propose \emph{video to behaviour} (ViBe), a new approach to learn models of road user behaviour from unlabelled raw video data of a traffic scene collected from a single,
monocular, initially uncalibrated camera with ordinary resolution. Our approach, illustrated in Figure~\ref{fig:vibe}, works by calibrating the camera (using available satellite images), detecting the relevant objects, and tracking them through time.
Each trajectory, together with the static and dynamic context of that road user at each moment in time, is then fed as a demonstration to our LfD system, which can learn robust behaviour models for road users.
The resulting models are then used to populate a simulation of the scene built using the Unity game engine.

The contributions of this paper are two-fold:
First, we present a vision pipeline that can track different road users and map their tracked trajectories to 3D space and is competitive with the state-of-the art approaches for image space tracking.
Second, we extend \emph{generative adversarial imitation learning} (GAIL) \cite{ho2016generative},
a state-of-the-art LfD method, with a novel curriculum-based training regime that enables our agents to gradually learn to mimic temporally extended expert demonstrations and successfully generalise to unseen situations.
We evaluate our method against several baselines, including \emph{behavioural cloning} (BC) and
state-of-the-art variants of GAIL. Using a number of metrics, we show that our
method can better imitate the observed demonstrations and results in more stable learning.

\vspace{-3pt}
\section{Related Work}
\label{literature}

\subsection{Computer Vision}
\vspace{-3pt}
In recent years, neural network approaches have significantly advanced the state of the art in computer vision tasks such as classification \cite{He2016DeepRL} and object detection \cite{Detectron2018}. Object detection is usually performed using region-based object detectors such as
 Fast R-CNN \cite{girshick15fastrcnn}, Faster R-CNN \cite{ren2015faster}, or Mask R-CNN \cite{he2017maskrcnn}. Such methods are usually slower but more accurate than single-object detectors such as SSD \cite{Liu16},
YOLO \cite{DBLP:journals/corr/abs-1804-02767}, RetinaNet \cite{DBLP:conf/iccv/LinGGHD17}, and hence more appropriate for the application considered here.

When tracking multiple objects, \emph{tracking by detection}, in which objects are first detected, then associated into tracks, is usually preferred. State-of-the art
tracking methods employ deep features \cite{Wojke2017simple,Wojke2018deep}
often generated by Siamese networks \cite{LealTaix2016LearningBT,Bertinetto2016FullyConvolutionalSN} alongside image space motion models \cite{Bewley2016_sort} and \emph{intersection over union} (IOU) trackers \cite{1517Bochinski2017}.

Our work employs a number of techniques for robust detection and tracking. However,
unlike most vision pipelines, ours maps detections to 3D space, and makes extensive use of 3D information while tracking. Recent work \cite{chang2018video} explores
a similar application and uses the resulting 3D trajectories to estimate car velocities and detect traffic anomalies. By contrast, we use the trajectories as input to LfD.

\subsection{Learning from Demonstration}
ViBe's LfD component extends GAIL \cite{ho2016generative} which is inspired by inverse reinforcement learning \cite{ng2000algorithms,ziebart2008maximum,abbeel2004apprenticeship} and is discussed further in Section~\ref{sec:background}. A wide range of LfD techniques have been developed using supervised, unsupervised, or reinforcement learning \cite{osa2018algorithmic}.
However, most methods \cite{argall2009survey, duan2017one, finn2016guided}, even when using raw video as sensory input \cite{liu2017imitation}, rely on either artificially generated demonstrations or those collected by specially deployed sensors, limiting their application in realistic domains.

By contrast, ViBe leverages demonstrations of behaviour that was occurring naturally. The same idea has been used to imitate basketball teams \cite{zhan2018generative}, predict taxi driver behaviour \cite{ziebart2008navigate},
and control complex animations \cite{peng2018deepmimic}. However,
all these methods still rely on sensors (or manual labelling) that provide ground truth information about the observed demonstrations, whereas ViBe extracts trajectories directly from raw, unlabelled videos; the satellite  images used for calibration are the only external input required.

Related to ViBe are several existing LfD methods that learn road and pedestrian behaviour \cite{vasquez2014inverse, li2017infogail, lee2017desire, henry2010learning}. Most relevant is learning highway merging behaviour \cite{kuefler2017imitating, bhattacharyya2018multi} from NGSIM \cite{colyar2007us}, a publicly available dataset of vehicle trajectories. However, these methods again rely on manual labelling, synthetic data or specialised
equipment to obtain the trajectories, while ViBe learns from raw, unlabelled videos of behaviour.

Recent work proposed a method that can learn to play ATARI games by observing YouTube videos \cite{aytar2018playing}. Like ViBe, this method leverages raw videos, and existing publicly available data. However, it trains only a single agent operating in 2D space, whereas ViBe learns to control multiple interacting agents in 3D space.

Concurrently to our work, Peng et al.~\cite{peng2018sfv} proposed a similar approach in the context of character animation. An off-the-shelf vision module extracts 3D poses from unstructured YouTube videos of single agents performing acrobatic motions. A simple LfD approach then rewards behaviour that matches waypoints in individual demonstrations. By contrast, we consider a more challenging setting with multiple agents, occlusions, and complex interactions between agents.  Consequently, behaviour detection, reconstruction, and imitation are more difficult. In particular, interactions between agents preclude a waypoint-matching approach, as there is no unique set of waypoints for an agent to match that would be robust to changes in other agents' behaviour.

\vspace{-5pt}
\section{Background \label{sec:background}}

To realistically model the traffic environment of an autonomous vehicle, we need to simulate multiple agents interacting in the same environment.  Unfortunately, due to the large number of road users that may populate a traffic scenario, learning a centralised policy to control all agents simultaneously is impractical.
The size of the joint action space of such a policy grows exponentially in the number of agents, leading to poor scalability in learning.  Furthermore, it is crucial to model variable numbers of agents (e.g., cars routinely enter and leave an intersection), to which such \emph{joint policies} are poorly suited (each agent typically has a fixed agent index).

To this end, we take an approach similar to that of \emph{independent $Q$-learning} (IQL) \cite{tan1993multi},
where each agent learns its own policy, conditioned only on its own observations.
The other actors are effectively treated as part of the environment.  We can then treat the problem as one of single-agent learning and share the parameters of the policy across multiple agents.
Parameter sharing \cite{gupta2017cooperative} avoids the exponential growth of the joint action space and elegantly handles variable numbers of agents. It also avoids instabilities associated with decentralised learning by essentially performing centralised learning with only one policy.

We model the problem as a \emph{Markov decision process} (MDP).
The MDP is defined by the tuple $(\mathcal{S}, \mathcal{A}, P, \mathit{R})$. $\mathcal{S}$ represents the set of environment states,
$\mathcal{A}$ the set of actions, $P(s_{t+1}|s_{t},a_{t})$ the transition function, and $\mathit{R(s_{t},a_{t})}$ the reward function.
We use $\pi$ for the stochastic policy learnt by our agent and $\pi_E$ for the expert policy which we can access only through a dataset $\mathcal{D}_{E}$.
The agent does not have access to $\mathit{R(s_{t},a_{t})}$ and instead must mimic the expert's demonstrated behaviour.
Given a dataset $\mathcal{D}_{E}$, we denote sample trajectories as $\tau^E$.
They consist of sequences of observation-action pairs generated by the expert $\tau^E = \{(s^E_{_1}, a^E_{_1}), \dots, (s^E_{_T}, a^E_{_T})\}$.
Similarly, we denote trajectories generated by our agent as $\tau = \{(s_{_1}, a_{_1}), \dots, (s_{_T}, a_{_T})\}$. In our case, $\mathcal{D}_{E}$ is obtained from raw videos, via the process described in Section~\ref{pipeline}.

The simplest form of LfD is \emph{behavioural cloning} (BC) \cite{pomerleau1989alvinn, ross2011reduction},
which trains a regressor (i.e., a policy) to replicate the expert's behaviour given an expert state. BC works well for states covered by the training distribution but generalises poorly due to compounding errors in the actions,
a problem also referred to as \emph{covariate shift} \cite{ross2010efficient}. By contrast, GAIL \cite{ho2016generative} avoids this by learning via interaction with the environment, similar to \emph{inverse reinforcement learning} \cite{ng2000algorithms} methods.

GAIL aims to learn a deep neural network policy $\pi_{\theta}$ that cannot be distinguished from the expert policy $\pi_{E}$.
To this end, it trains a \emph{discriminator} $D_{\phi}$, also a deep neural network, to distinguish between state-action pairs coming from expert and agent (a process similar to GANs \cite{goodfellow2014generative}). GAIL optimises $\pi_{\theta}$ to make it difficult for the discriminator to make this distinction. Formally, the GAIL objective is:
\begin{align*}
  \min_{\theta} \max_{\phi} \expectation_{\pi_{\theta}}
    \left[ \log(D_\phi(s, a)) \right] +
  \expectation_{\pi_{E}}  \left[ \log(1 - D_\phi(s^{E},a^{E})) \right].
\end{align*}
Here, $D_{\phi}$ outputs the probability that $(s, a)$ originated from $\pi_{\theta}$.
As the agent interacts with the environment using $\pi_{\theta}$, $(s, a)$ pairs are collected and used to train $D_{\phi}$.
Then, GAIL alternates between a gradient step on $\phi$ to increase the objective function with respect to $D$, and an RL step on $\theta$ to decrease it with respect to $\pi$.
Optimisation of $\pi$ can be done with any RL algorithm using a reward function of the form $R(s,a)= -\log(D_{\phi}(s,a))$.
Typically, GAIL uses policy gradient methods that approximate the gradient with Monte Carlo rollouts \cite{schulman2015trust} or a critic \cite{schulman2017proximal}.
Optimisation of $D_\phi$ minimises a cross entropy loss function.

Early in training, the state-action pairs visited by the policy are quite different from those in the demonstrations, which can yield unreliable and sparse rewards from $D_{\phi}$, making it difficult to learn $\pi_{E}$.
We show how we address this problem by introducing a novel curriculum in Section~\ref{algorithm}.

In multi-agent situations, GAIL agents trained in a single-agent setting can fail to generalise to multi-agent settings \cite{bhattacharyya2018multi}.
PS-GAIL \cite{bhattacharyya2018multi} is an extension to GAIL that addresses this issue by
gradually increasing the number of agents controlled by the policy during training.
We compare to PS-GAIL experimentally in Section~\ref{experiments}. However, it is complementary to the Horizon GAIL method we propose in Section~\ref{algorithm} and future work can focus on using them in conjunction.

\vspace{-5pt}
\section{ViBe: Video to Behaviour}
\label{pipeline}
In this section, we describe ViBe, which learns road behaviour policies from raw traffic camera videos (see Figure~\ref{fig:vibe}). We first describe how trajectories are extracted from these videos. We then describe how they are used to create a simulation of the scene. Finally, we detail how the trajectories and the simulator are used to learn realistic road behaviour policies via our novel LfD approach.

\vspace{-2pt}
\subsection{Extracting Demonstrations}
\label{cv}
\vspace{-2pt}
\begin{figure}[t!]
  \centering
  \includegraphics[width=0.3\textwidth]{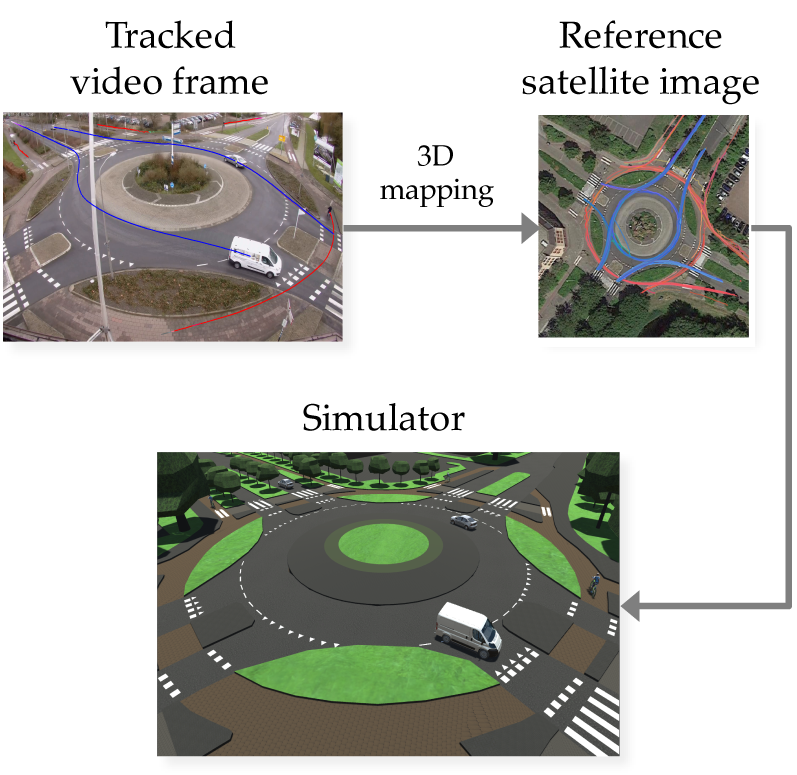}
  \vspace{2pt}
  \caption{Example of how ViBe's vision module tracks cars (blue) and pedestrians (red). The tracks are projected to 3D space using a reference satellite image from Google Maps. The tracks are played back in a simulation of the scene developed in Unity. }
  \label{fig:cv_fig}
  \vspace{-15pt}
\end{figure}

This section describes our vision pipeline, whose main steps are \emph{detection}, \emph{calibration}, and \emph{tracking}. Our detector uses the bounding box output of a pre-trained model of Mask R-CNN \cite{he2017maskrcnn} \cite{Detectron2018} based on the ResNet-101 \cite{He2016DeepRL} architecture, pre-trained on the COCO dataset \cite{Lin2014MicrosoftCC}.  Since we are only interested in the traffic information, we remove all classes except car, bus, truck, pedestrian, and bicycle.

The next step is calibration. As traffic cameras tend to have a large field of view, the camera images tend to be highly distorted. As we do not have access to the cameras, we are unable to calibrate the camera using traditional methods (e.g., using a checkboard) \cite{Bradski:2008:LOE:1461412}. Instead, we obtain a top-down satellite image of the scene from Google Maps and add landmark points to both camera and satellite images. We then undistort the camera image and use the landmark points to calculate the camera matrix. Given the camera calibration we map the detected bounding boxes into 3D by assuming that the detected object is a fixed height above the ground, with the height depending on its class.

The final step is tracking multiple objects in unstructured environments. Our multiple object tracking (MOT) module is similar to that of Deep SORT \cite{Wojke2018deep}, which makes use of an \emph{appearance model} to make associations. For each
scene, we train an appearance model using a \emph{Siamese network} (SN) \cite{LealTaix2016LearningBT}. We first run our object detector over the whole video, followed by an IOU tracker. This yields short tracks that we call
\emph{tracklets}. Objects in the same tracklets form positive pairs, and objects from different tracklets form negative pairs used to train the SN. To avoid the possibility of similar objects
appearing in negative pairs, we form these pairs using tracklets with a large temporal difference. The SN is trained using a cosine distance metric and a contrastive loss.

Our MOT pipeline then processes the detected objects through several steps. Track initialisation occurs when a simple IOU tracker associates more than five consecutive detections. The initialised track is mapped to 3D space, where a Kalman filter predicts the next position of the object. Next, objects in the next frame within the vicinity of this prediction are compared with the current track using the features generated by the SN. An association is made if this comparison yields a cosine distance in the feature space below a certain threshold.
If no such association is made, the tracker attempts to associate detections using IOU. If association still fails, a final attempt is made using nearest neighbour association in 3D. Figure~\ref{fig:cv_fig} shows an example output of our tracking pipeline in both 2D and 3D space.

\vspace{-5pt}
\subsection{Simulation}
\label{sim}
\vspace{-3pt}

Our vision pipeline outputs timestamped trajectories of different road users. However, a simulator also requires a reliable representation of the static elements of the scene such as pavements and zebra crossings.
To this end, we use Google Maps as a reference to build a simulation of the scene in Unity. Building the static elements of the simulation is straightforward and significantly easier than realistically
modeling the dynamic elements of the scene. In this paper, we simulate a roundabout intersection in Purmerend, a city in the Netherlands that provided the traffic video data used in our experiments. Figure~\ref{fig:cv_fig} shows how the scene with some tracks from our vision pipeline is recreated in our simulator.

Section~\ref{algorithm} describes our LfD approach, which requires a state representation for the agent. Our simulator generates such a representation based on both the static and dynamic context.
Pseudo-LiDAR readings, similar to those in \cite{bhattacharyya2018multi}, are used to represent
different aspects of the static (e.g., zebra crossings and roads) and dynamic (e.g., distance and velocity of other agents) context of the agent. In addition, we provide information such as the agent's heading, distance from goal, and velocity.
Our simulator uses a simple linear motion model, which we found sufficient for learning, though in the future individual motion models for each road entity could be considered.

Given a start frame in the dataset, our simulator plays back
tracked trajectories from that frame onwards, produces observations, and accepts actions from agents controlled by
neural network policies.
In other words, it provides exactly the environment needed to both
perform LfD on the extracted trajectories and evaluate the resulting learnt
policies.

\subsection{Learning}
\label{algorithm}
\vspace{-3pt}
Given the trajectories extracted by the vision processing from Section~\ref{cv}, ViBe  uses the simulator from Section~\ref{sim} to learn a policy that matches those trajectories. Learning is based on GAIL, which leverages the simulator to train the agent's behaviour for states beyond those in the demonstrations, avoiding the compounding errors of BC.  However, in the original GAIL method, this interaction with the simulator means that the agent has control over the visited states from the beginning of learning. Consequently, it is likely to take bad actions that lead it to undesirable states, far from those visited by the expert, which in turn yields sparse rewards from the discriminator and slow agent learning.

To address this problem, we propose \emph{Horizon GAIL}, which, like BC, bootstraps learning from the expert's states, in this case to ensure a reliable reward signal from the discriminator.  To prevent compounding errors, we use a novel horizon curriculum that slowly increases the number of timesteps for which the agent interacts with the simulator. Thus, only at the end of the curriculum does the agent have the full control over visited states that the original GAIL agent has from the beginning.  This curriculum also encourages the discriminator to learn better representations early on.

\begin{figure}
  \centering
  \includegraphics[width=0.35\textwidth]{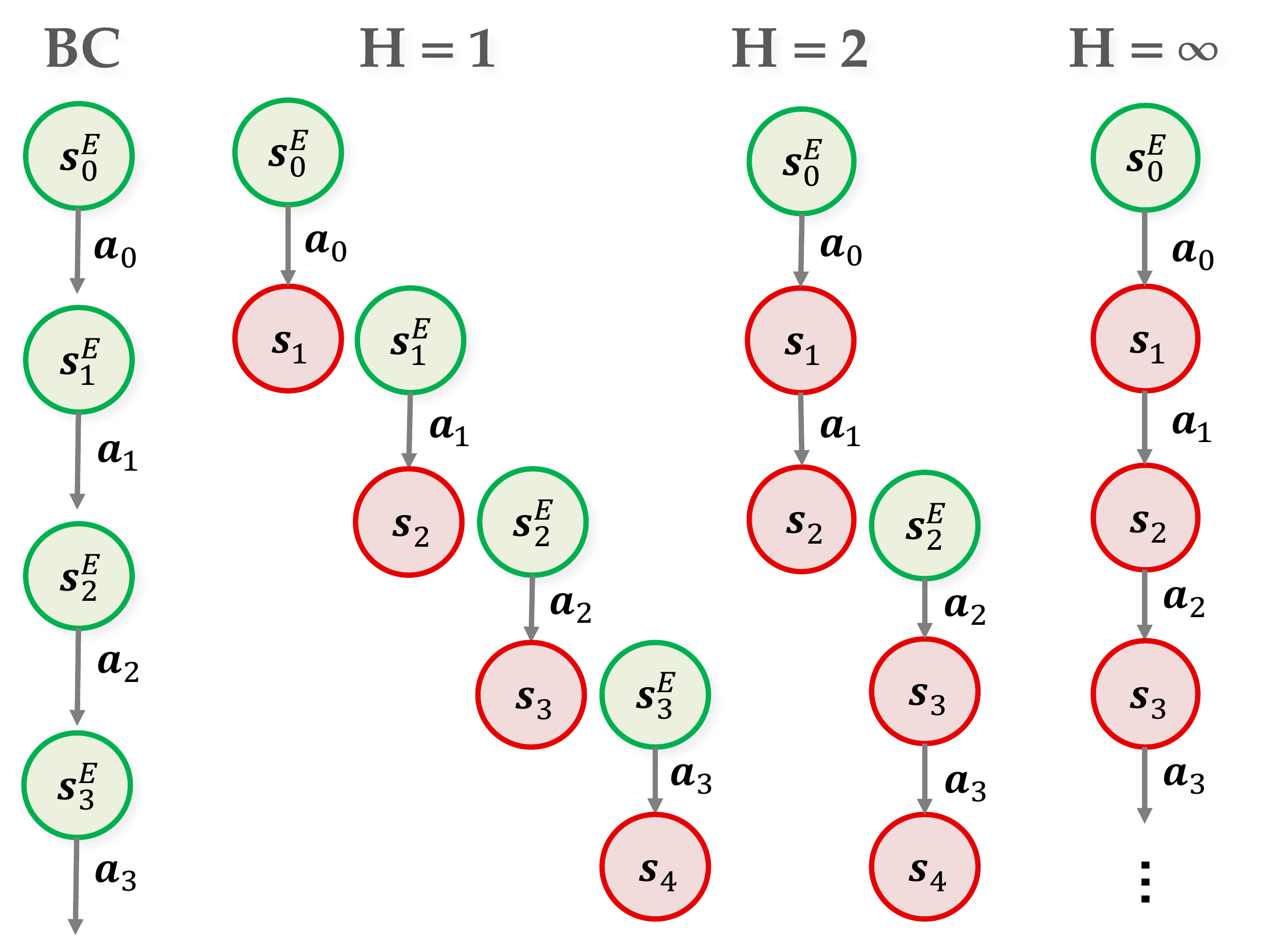}
  \vspace{-5pt}
  \caption{Schematic of Horizon GAIL for different values of the horizon $H$, compared to BC. Green circles indicate bootstrapped expert states, red circles correspond to states that the agent encounters after acting in the environment. When $H=\infty$, Horizon GAIL matches original GAIL.}
  \label{fig:horizon}
  \vspace{-20pt}
\end{figure}

In each episode, the agent is initialised from a random expert state, $s_t^E$ and must act for $H$ steps, where $H$ is the length of the horizon. Once the horizon is reached, the simulation ends but the episode is not considered terminated.  Instead, Horizon GAIL uses an actor-critic approach, with the actor following a gradient estimated from an $n$-step return, with $n=H$, bootstrapping from a critic $V_{\psi}$ when the horizon is reached.  This prevents the agent from learning myopic behaviour when $H$ is small. Hence, while GAIL is agnostic about the policy gradient method it uses, Horizon GAIL requires a critic in order to bootstrap beyond the simulated horizon.

When $H=1$, Horizon GAIL is similar to BC. In fact, pre-training GAIL with BC is known to be beneficial \cite{kuefler2017imitating, li2017infogail,wang2017robust, song2018multi}. However, even with $H=1$, a crucial difference remains (see Figure~\ref{fig:horizon}).
BC does not interact with a simulator, as the agent simply learns to predict the expert's action given its state. By contrast, when $H=1$, the Horizon GAIL agent's action is fed back into the simulator, which generates $s_{t+1}$ and the policy gradient estimate bootstraps with $V_\psi(s_{t+1})$. When $H=2$, the agent, initialised from $s_t^E$, acts for two steps in the simulator before being terminated. $H$ is increased during training according to a schedule. When $H=\infty$, Horizon GAIL is equivalent to GAIL.

Gradually moving from single step state-action pairs to more difficult multi-step trajectories helps stabilise learning.
It allows the generator and discriminator to jointly learn to generalise to longer sequences of behaviour and match the expert data more closely while ensuring the discriminator does not collapse early in training. We found that Horizon GAIL was critical to successfully reproduce naturalistic behaviour in our complex traffic intersection problem, as we show in Section~\ref{results}.

\section{Experimental results}
\label{experiments}
\vspace{-3pt}

We evaluate ViBe on a complex multi-agent traffic scene involving a roundabout in Purmerend (Section~\ref{sim}). The input data consists of 850 minutes of video at 15 Hz from the traffic camera observing the roundabout.
Our vision pipeline identifies all the agents in the scene (e.g., cars, pedestrians and cyclists), and tracks their trajectories through time, resulting in around 10000 car trajectories. Before
any learning, these trajectories are filtered and pruned. Specifically, any trajectories that result in collisions or very large velocities are considered artifacts of the tracking process and are not used during training.
We split the resulting dataset into training, validation, and test sets such that there is no temporal overlap, i.e., no test trajectories occur at the same time as training trajectories.
The validation set is used to tune hyperparameters and choose the best performing model (for all baselines) in evaluation.
As discussed in Section~ \ref{sim}, we can use our simulator to play back these trajectories at any point in time (see Figure~\ref{fig:vibe}).

When training with Horizon GAIL, in each episode the agent is initialised at a point sampled from an expert trajectory. The sampled point determines the full initial state of the simulator, including position, velocity, and heading of all agents in the scene. We use our policy to simulate the agent for $H$ steps. The agent is also assigned a goal corresponding to the last state of the expert trajectory.
The episode terminates if the agent collides with an object or another agent, or reaches its goal.

We compare Horizon GAIL to a number of  baselines: BC, GAIL \cite{ho2016generative} and PS-GAIL \cite{bhattacharyya2018multi}, using the same dataset and observation and action spaces to train all methods. We show results using the best hyperparameters we found after tuning them separately for each method.
\begin{table}[b!]
  \vspace{-5pt}
  \centering
  \begingroup
  \setlength{\tabcolsep}{10pt} 
  \renewcommand{\arraystretch}{1} 
    \caption{Comparison of ViBe vision module to baseline trackers}
  \begin{tabular}{lllll}
  \toprule
               & \textbf{NT}          & \textbf{IDF1}            & \textbf{IDP}             & \textbf{IDR}             \\ \toprule
               IOU         & 400         & 51.1\%          & 50.3\%          & 51.8\%          \\
  Deep SORT & 129        & 68.1\%          & 66.6\%          & 69.7\%          \\
  \textbf{ViBe}  & \textbf{97} & \textbf{70.5\%} & \textbf{68.1\%} & \textbf{73.1\%} \\
  \bottomrule
  \end{tabular}
  \label{tab:cv_res}
  \endgroup
\end{table}

Policies, $\pi_{\theta}$, take as input 64 dimensional pseudo-LiDAR observations with a field of view of $2\pi$ radians, generated by our simulator as described in Section~\ref{sim}.
These LiDAR observations are stacked together and processed in two layers of 1x1 convolutions of 15 and 3 channels respectively. These convolutions act as channel mixing operations but maintain the spatial information
of the original signal. The output then passes through a series of fully connected layers and is concatenated with the agent's orientation, distance
from the goal, and a one-hot encoding of the target roundabout exit. The network outputs displacements in Cartesian coordinates, used by the simulator to update the agent's location.

\begin{figure*}[t!]
  \centering
  \includegraphics[width=0.80\textwidth]{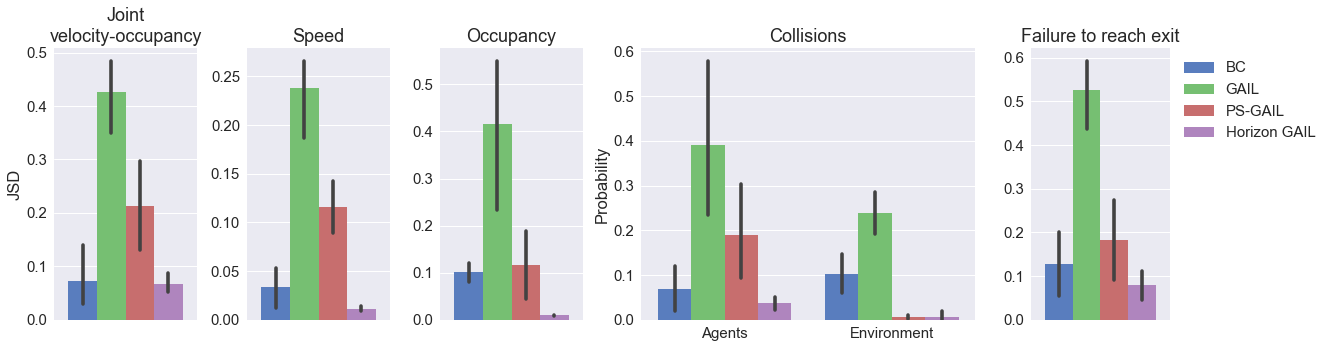}
  \vspace{-10pt}
  \caption{Results of evaluation across 4 independent 4000 timesteps of multi-agent simulations across different metrics: Jensen-Shannon divergence between joint velocity-occupancy, speed and occupancy distributions of ground truth and simulated agents. The collision probability, either with other agents or the environment. Probability of failing to reach the correct exit.}
  \vspace{-5pt}
  \label{fig:metrics}
\end{figure*}

We use
identical core architectures for the discriminator $D_{\phi}$ and value function $V_{\psi}$. Unlike \cite{bhattacharyya2018multi}, we do not represent the policy using a recurrent neural network, as we found that a feedforward network worked well in practice.

We train $\pi_{\theta}$ with \emph{proximal policy optimisation} (PPO) with a clipping objective \cite{schulman2017proximal}, an actor-critic method known to perform well for long-horizon problems \cite{OpenAIdota}.
We train each model for 5000 epochs, each containing 1024 environment interactions. For Horizon GAIL, the horizon schedule starts with $H=1$ and increments by 1 every 100 epochs. However, performance is quite robust to this hyperparameter: varying the schedule from 50 to 200 epochs did not create any significant performance differences.

\vspace{-3pt}
\subsection{Performance Metrics}
\label{metrics}
\vspace{-3pt}

\begin{figure*}[t!]
  \centering
  \includegraphics[width=0.75\textwidth]{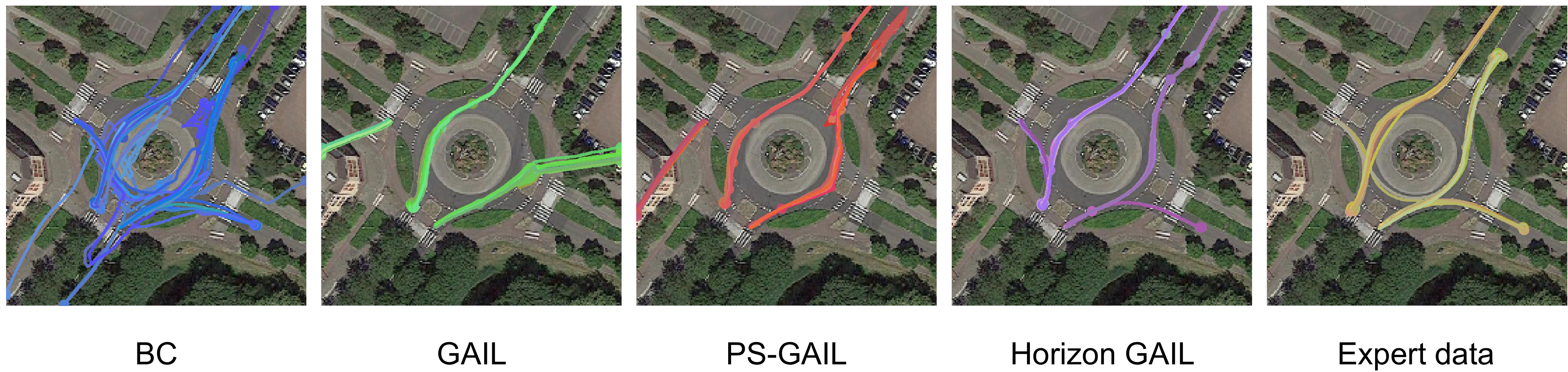}
  \vspace{-10pt}
  \caption{Top views of the trajectories taken by the agents, when trying to replicate the expert trajectories shown on the right-most column. These trajectories are produced across 4000 timesteps of multi-agent simulation.}
  \vspace{-15pt}
  \label{fig:traces}
\end{figure*}

To evaluate the ViBe vision module, we measure the reliability of the tracks it generates using the metrics introduced by Ristani et al.\ \cite{Ristani2016PerformanceMA}:
number of tracked trajectories (NT), identity F1 score (IDF1), identity precision (IDP) and identity recall (IDR). These metrics are suitable because they reflect the key qualities of reliably tracked trajectories.

To evaluate our policies, we chose a 4000 timestep window of the test data and simulated all the cars within that interval. These windows do not overlap for each evaluation run.
Pedestrians and other road users are played back from the dataset.
In contrast to training, during evaluation we do not terminate the agents upon collision, so as to assess how well each method models long term behaviour.

Unlike in reinforcement learning, where the true reward function is known, performance evaluation in LfD is not straightforward and typically no single metric suffices.
Several researchers have proposed metrics for LfD, which are often task specific \cite{kuefler2017imitating, shiarlis2016inverse, wang2017robust}.
We take a similar approach, using a suite of metrics, each comparing a different aspect of the generated behaviour to that of human behaviour.

During evaluation we record the positions and velocities of all simulated agents. Using kernel density estimation,
we estimate probability distributions for speed and 2D space occupancy (i.e., locations in 2D space) as well as a joint distribution of velocities and space occupancy. The same distributions are computed for the
ground truth data. We then measure the Jensen-Shannon divergence (JSD) between the data and the respective model generated distributions for these three quantities.
We also measure how often the simulated agents collide with objects or other agents in the environment, i.e., the collision rate. Finally, we measure how often the agents fail to reach their goal.
\begin{figure}[b!]
  \vspace{-15pt}
  \centering
  \includegraphics[width=0.43\textwidth]{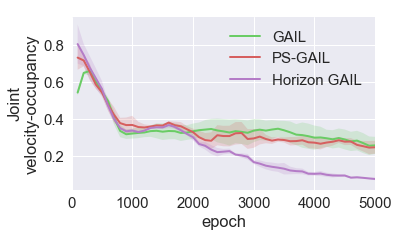}
  \vspace{-10pt}
  \caption{Progression of Joint velocity-occupancy JSD metric through training, indicating  difference in stability between our method (Horizon GAIL) and other GAIL baselines across 3 random seeds.}
  \label{fig:training_curve}
\end{figure}

\subsection{Results}
\label{results}

To validate the ViBe vision module, we manually label
43 trajectories from the dataset and then compare its performance against two baselines, a simple IOU \cite{1517Bochinski2017} tracker and Deep SORT \cite{Wojke2017simple,Wojke2018deep}, a state-of-the-art MOT pipeline. We replace Deep SORT's appearance model with our own, as it is specifically trained for this scene.

The results in Table~\ref{tab:cv_res} show that the ViBe vision module outperforms both baselines. In particular, ViBe's higher IDF1 score gives confidence that the trajectories
provided are of sufficient quality for LfD. The most substantial difference between
Deep SORT and ViBe is that ViBe performs Kalman filtering in 3D space, which likely explains the performance difference. Even for ViBe, the number of tracked trajectories~(NT) is substantially higher than ground truth (43).  However, this is not caused by false trajectories but merely by the tracker splitting a single trajectory into separate ones.
This in turn implies that ViBe produces longer tracks than the baseline methods.

The results of our LfD evaluation can be seen in the following figures: Figure~\ref{fig:metrics} shows performance with respect to the evaluation metrics discussed in Section~\ref{metrics} for 4 independent 4000 timesteps of multi-agent simulations. Figure~\ref{fig:traces} shows the trajectories generated by a single such simulation by each method. Horizon GAIL outperforms all baselines and produces trajectories that
more closely resemble the data. GAIL and PS-GAIL perform relatively poorly, failing to capture the data distribution. These results represent the best training epoch out of the 5000 performed during training,
as we observed that both baseline GAIL methods exhibit quite unstable training dynamics. Figure~\ref{fig:training_curve}, which plots the joint velocity-occupancy JSD metric across the training epochs for a multi-agent evaluation of 4000 timesteps across 3 random seeds, shows  that
Horizon GAIL is noticeably more stable. With respect to PS-GAIL, we observed that the curriculum parameter was relatively hard to tune. For example, adding agents too soon causes the discriminator to learn too quickly that these agents are not real.

Another notable observation is that BC performs well when compared to both baseline GAIL methods. This result can be attributed to the abundance of data available for training. From Figure~\ref{fig:traces} however we can see that
the qualitative performance of these policies is relatively poor when compared to Horizon GAIL. As expected, the BC baseline quickly diverges from plausible trajectories, as minor errors compound over time.
The long evaluation times exacerbate this effect.
Horizon GAIL avoids compounding error problems associated with BC through interaction with the environment. It also avoids unstable training related with GAIL through the gradually increasing horizon. This yields stable, plausible trajectories with fewer collisions than any other method.\footnote{Accompanying video: \url{https://youtu.be/3VK4tQTHeHc}.}
\vspace{-7pt}

\section{Conclusion}
\label{conclusion}
\vspace{-2pt}
This paper presented a novel method for learning from demonstration in the wild that can leverage abundance of freely available videos of natural behaviour.
In particular, we proposed ViBe, a new approach to learning models of road user behaviour from unlabelled raw video data of a traffic scene collected from a single, monocular, uncalibrated camera with ordinary resolution.  ViBe calibrates the camera, detects relevant objects, tracks them reliably through time, and uses the resulting trajectories to learn driver policies via a novel LfD method. The learned policies are finally deployed in a simulation of the scene developed using the Unity game engine. According to several metrics our LfD method exhibits better and more stable learning than baselines such as GAIL and BC.

\bibliographystyle{IEEEtran}
\bibliography{IEEEabrv,refs.bib}

\pagebreak
\appendix
\section{Appendix}
\label{appendix}
\vspace{15pt}
\subsection{Vision}
\vspace{2pt}
The vision pipeline discussed in Section \ref{cv} calibrates the camera, detects relevant objects, tracks them from one frame to the next, and provides our LfD system with demonstrations. Here we provide further details on the tracking system.

Tracking is performed in two consecutive stages using appearance features and IOU (intersection-over-union) described in Section~\ref{literature}. In the first stage, for a certain frame $N$ in the video, given existing tracks and detections generated from Mask R-CNN \cite{he2017maskrcnn}, an appearance cost matrix is computed, where each element is the distance between the features of each detection and each track. The feature distances between one detection in the $N$th frame and each detection in a track are calculated, and the minimum value is selected.
\vspace{10pt}
\begin{figure}[H]
  \centering
  \includegraphics[width=0.5\textwidth]{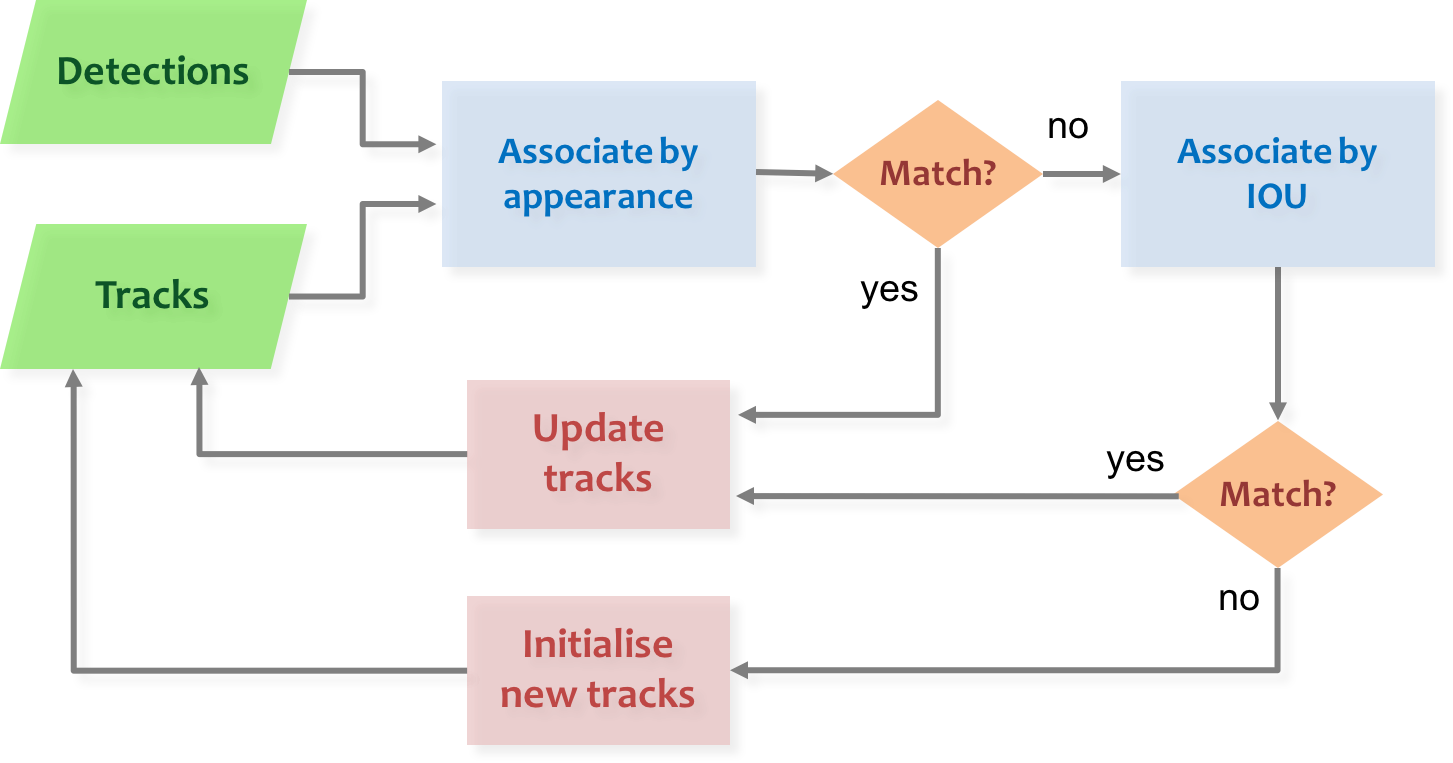}
  \caption{Schematic of ViBe's tracker:   Detections generated from Mask R-CNN are matched with tracks in two stages, using appearance features and IOU. Detections with no matches are initialised as new tracks.}
  \label{fig:tracking}
\end{figure}
\vspace{5pt}

For each track, the associated 3D Kalman filter predicts the next position of the track. We only consider detections for the next step to be valid if they fall within a fixed radius of the Kalman filter's prediction.
Additionally, we also only consider detections to be valid if they have a cosine distance in the Siamese network's feature space below a certain threshold compared to the current track.

These two gates are applied on the appearance cost matrix to generate a new matrix where each element outside the valid range is substituted by a large value. The Hungarian algorithm is then applied on this new cost matrix to provide matched tracks and detections.

The unmatched tracks and detections are then fed into the second stage. An IOU cost matrix is calculated between the unmatched detections and the last successful detection of unmatched tracks. Similarly, a threshold is applied to the appearance cost matrix between the unmatched detections and the unmatched tracks. The IOU matrix is modified according to the appearance gating. The Hungarian algorithm is applied and the result provides the matched detections and matched tracks. Unmatched detections are used to initialise new tracks.
Unmatched tracks are removed from the tracking process if the number of frames without detection association exceeds a threshold.
The tracks that have been matched in this frame given the 2 stages and the newly initialised tracks are then fed into the next frame, which repeats the above process.

Figure~\ref{fig:tracking} depicts this process. The finalised tracks are then played back in our simulator to provide the demonstration trajectories used by our LfD algorithm (see Section~\ref{sim}).

\vspace{10pt}
\subsection{LfD}
\vspace{3pt}
Here we provide some implementation details about our LfD approach. Given the trajectories extracted by our visual processing module, ViBe uses the simulator to learn a policy that imitates those trajectories using our novel Horizon GAIL method, described in Section~\ref{algorithm}. Algorithm~\ref{alg0} provides a complete overview of our training scheme.
\vspace{8pt}
\begin{algorithm}
  \caption{Horizon Curriculum for GAIL}
  \label{alg0}
\begin{algorithmic}
  \STATE Initialise policy $\pi_{\theta}$, discriminator $D_{\phi}$, expert demonstrations $\mathcal{D}_{E}$
  \FOR{$h = 1 \ldots T$}
    \STATE Sample expert trajectory: $\tau_E \sim \mathcal{D}_{E}$
    \FOR{$t = 0, h, 2h, \ldots, T-h$}
      \STATE Use expert observation $s^E_t$ to initialise the agent and initialise the environment to its corresponding state at time $t$
      \STATE Sample an agent's trajectory of length $h$: $\tau \sim \pi_{\theta_i}$
    \ENDFOR
    \STATE Sample $M$ observation-action pairs $\chi \sim \tau$ and $M$ pairs $\chi_E \sim \tau_E$
    \STATE Update the discriminator parameters from $\phi_i$ to $\phi_{i+1}$ with the gradient:
    \begin{multline*}
      \expectation_{(s_m, a_m) \in \chi}
        \left[ \nabla_\phi  \log(D_\phi(s_m, a_m)) \right]
      \\
      + \expectation_{(s^E_m, a^E_m) \in \chi_E}  \left[ \nabla_\phi  \log(1 - D_\phi(s^{E}_m,a^{E}_m)) \right]
    \end{multline*}
    \STATE Compute reward $\forall (s_m, a_m) \in \chi$ using the discriminator: $r_m = -\log(D_{\phi_{i+1}}(s_m, a_m))$
    \STATE Take a policy step from $\theta_i$ to $\theta_{i+1}$, with any policy optimisation method
  \ENDFOR
\end{algorithmic}
\end{algorithm}
\vspace{7pt}

Horizon GAIL bootstraps learning from the expert's states, $s^E_t$, to ensure a reliable reward signal from the discriminator and helps stabilise learning by using a novel horizon curriculum that slowly increases the number of timesteps, horizon $h$, for which the agent interacts with the simulator.

Gradually moving from single step state-action pairs to more difficult multi-step trajectories allows the generator and discriminator to jointly learn to generalise to longer sequences of behaviour and match the expert data more closely while ensuring the discriminator does not collapse early in training.

\newpage

\begin{figure}[H]
    \centering
    \includegraphics[width=0.47\textwidth]{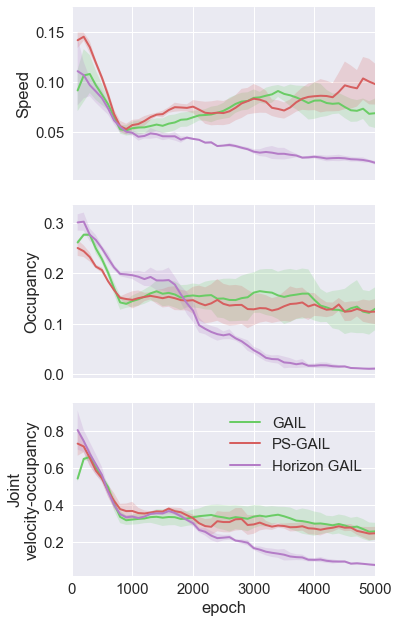}
    \caption{Progression of speed, occupancy and joint velocity-occupancy JSD metrics throughout training for our method (Horizon GAIL) and other GAIL baselines across three random seeds.}
    \label{fig:training_curves_all}
  \end{figure}

The results of our LfD evaluation can be seen Section~\ref{results}. Here we extend Figure~\ref{fig:training_curve} to also include speed and occupancy JSD metrics (described in Section~\ref{metrics}) across the training epochs for a multi-agent evaluation of 4000 timesteps of the held-out data across three random seeds (see Figure~\ref{fig:training_curves_all}). Throughout training, Horizon GAIL exhibits noticeably more stable behaviour across these metrics.

\end{document}